\documentclass[sigconf, screen, nonacm]{acmart}

\usepackage{booktabs}
\usepackage{catchfile}
\usepackage{tikz}
\usetikzlibrary{arrows.meta,decorations.pathmorphing,calc}

\newcommand{\tbd}{\textcolor{red}{\textbf{?}}}
\newcommand{\site}{\url{https://wrong.cuyor.com}}
\IfFileExists{results/macros.tex}{\providecommand{\tbd}{\textcolor{red}{\textbf{?}}}
\newcommand{\modelMB}{834} %
\newcommand{\negPct}{30\%} %
\newcommand{\tokenCap}{50} %
\newcommand{\corrPct}{10\%} %
\newcommand{\verifyN}{200} %
\newcommand{\verifyAgree}{99.0\%} %
\newcommand{\benchTimed}{195} %
\newcommand{\benchWarmup}{5} %
\newcommand{\citedEm}{0.875} %
\newcommand{\citedSilent}{0.045} %
\newcommand{\citedRefused}{0.034} %
\newcommand{\citedNoParse}{0.046} %
\newcommand{\citedSeeds}{5} %
\newcommand{\nTest}{1{,}357} %
\newcommand{\brEm}{0.882} %
\newcommand{\brSilent}{0.036} %
\newcommand{\brSilentPct}{3.6\%} %
\newcommand{\brCorrect}{1{,}197} %
\newcommand{\brDetectable}{111} %
\newcommand{\brSilentN}{49} %
\newcommand{\brMedianS}{0.85\,s} %
\newcommand{\brRefused}{27} %
\newcommand{\brRefusedRate}{0.020} %
\newcommand{\brNoParse}{0.062} %
\newcommand{\brMalformed}{84} %
\newcommand{\brRestartNote}{During the run the WebAssembly worker failed 2 times; the page reloaded the model and retried the affected orders.} %
\newcommand{\srvAgree}{99.0\%} %
\newcommand{\srvDiffN}{13} %
\newcommand{\emRelation}{reaches the same exact match (0.882)} %
\newcommand{\probeN}{14} %
\newcommand{\probeRefused}{14} %
\newcommand{\parityN}{507{,}618} %
\newcommand{\parityDisagree}{0} %
\newcommand{\netOrders}{10} %
\newcommand{\netChars}{629} %
\newcommand{\netRequests}{0} %
\newcommand{\netBytes}{0} %
\newcommand{\decOutputs}{84{,}134} %
\newcommand{\decDisagree}{0} %
\newcommand{\nDevices}{3} %
\newcommand{\latencyReading}{On the MacBook Air (Apple M3, 16\,GB) in Chrome, the median order takes 383\,ms and nine in ten finish within 517\,ms, inside the 1\,s limit. On the MacBook (Apple M1, 8\,GB) in Chrome, the median order takes 600\,ms and nine in ten finish within 830\,ms, inside the 1\,s limit. On the Windows laptop (Intel Core i7-1165G7, 16\,GB) in Edge, the median order takes 2.99\,s and nine in ten finish within 4.56\,s. Over all devices, 3 orders needed a reload of the WebAssembly worker and are left out of the times.} %
\newcommand{\latencyBreakdown}{Before the first token the model reads the prompt: 176\,ms on the M3 MacBook Air, 267\,ms on the M1 MacBook and 1.88\,s on the Windows laptop (45--65\% of each order); the ticket then streams at 50, 31 and 10 tokens per second, respectively, for about 12 tokens per ticket. A first visit takes 40--45\,s on every device, of which 36--38\,s is the 834\,MB download.} %
\newcommand{\fidelityReading}{The 13 orders whose browser output differs are long ones: a median of 25 words against 13 over the test set, and 7 of them hold three or more items, against 8\% overall. Their effect on accuracy cancels out: 2 correct server tickets went wrong and 2 wrong ones became correct. Of the server's silent errors among them, 4 no longer parse in the browser and 1 new silent error appeared, so the tab is, if anything, louder.} %
}{}
\providecommand{\nTest}{1{,}357}
\IfFileExists{results/latency_rows.tex}{%
  \newcommand{\latencyRows}{    M3 Air, 16 GB, Chrome 157 & 383\,ms & 517\,ms & 0\% & 100\% & 42\,s \\ %
    M1 MacBook, 8 GB, Chrome 156 & 600\,ms & 830\,ms & 0\% & 98\% & 40\,s \\ %
    Dell i7-1165G7, 16 GB, Edge 156 & 2.99\,s & 4.56\,s & 0\% & 0\% & 45\,s \\ %
}}{%
  \def\latencyRows{%
    M1 Air 8\,GB, Chrome & \tbd & \tbd & \tbd & \tbd & \tbd \\
    Computer 2, Chrome   & \tbd & \tbd & \tbd & \tbd & \tbd \\
    Computer 3, Chrome   & \tbd & \tbd & \tbd & \tbd & \tbd \\}}
\newcommand{\code}[1]{\texttt{#1}}

\setcopyright{none}
\acmConference[WWW Companion '27]{Companion Proceedings of the ACM Web Conference 2027}{May 10--14, 2027}{Dublin, Ireland}
\acmBooktitle{Companion Proceedings of the ACM Web Conference 2027 (WWW Companion '27), May 10--14, 2027, Dublin, Ireland}

\begin{document}

\title{Clean Tickets, Wrong Orders}
\subtitle{Catching Silent Errors with a Language Model in Your Browser Tab}

\author{Umar Ahmed}
\orcid{0009-0003-6040-6328}
\affiliation{\institution{ITMO University}\city{Saint Petersburg}\country{Russia}}
\email{uakhmed@itmo.ru}

\author{Maxim Dunaev}
\orcid{0009-0009-2988-9839}
\affiliation{\institution{ITMO University}\city{Saint Petersburg}\country{Russia}}
\email{mvdunaev@itmo.ru}

\author{Mohammad Nour Al Awad}
\orcid{0009-0008-5078-9800}
\affiliation{\institution{ITMO University}\city{Saint Petersburg}\country{Russia}}
\email{MohammadNourAlAwad@itmo.ru}

\author{Sergey Ivanov}
\orcid{0000-0002-1128-2942}
\affiliation{\institution{ITMO University}\city{Saint Petersburg}\country{Russia}}
\email{svivanov@itmo.ru}

\author{Ivan Khodnenko}
\orcid{0009-0003-7787-7126}
\affiliation{\institution{ITMO University}\city{Saint Petersburg}\country{Russia}}
\email{ivan.khodnenko@itmo.ru}

\renewcommand{\shortauthors}{Ahmed et al.}

\begin{abstract}
A small language model that turns a typed order into structure can fail in two
ways. It can print something broken, which any system rejects. Or it can print
a clean, well formed order that nobody placed. Our companion work calls the
second kind a \emph{silent error}, and it is the dangerous one. We present a web demo in which
a fine-tuned 0.8B parameter model %
runs entirely inside the visitor's browser tab, through llama.cpp compiled to
WebAssembly. No order text leaves the device. The page parses orders as the
visitor types, scores every ticket with the PIZZA benchmark's own matcher ported
to TypeScript, refuses anything off the menu with ``\code{?}'' so the model can
face the open web, and lets visitors hunt for silent errors or try to cause one.
Run in the browser on all \nTest{} PIZZA test orders, the deployed model writes
the same output as the server evaluation for \srvAgree{} of them and
\emRelation; we also report its latency on \nDevices{} consumer laptops.
Demo: \site.
\end{abstract}


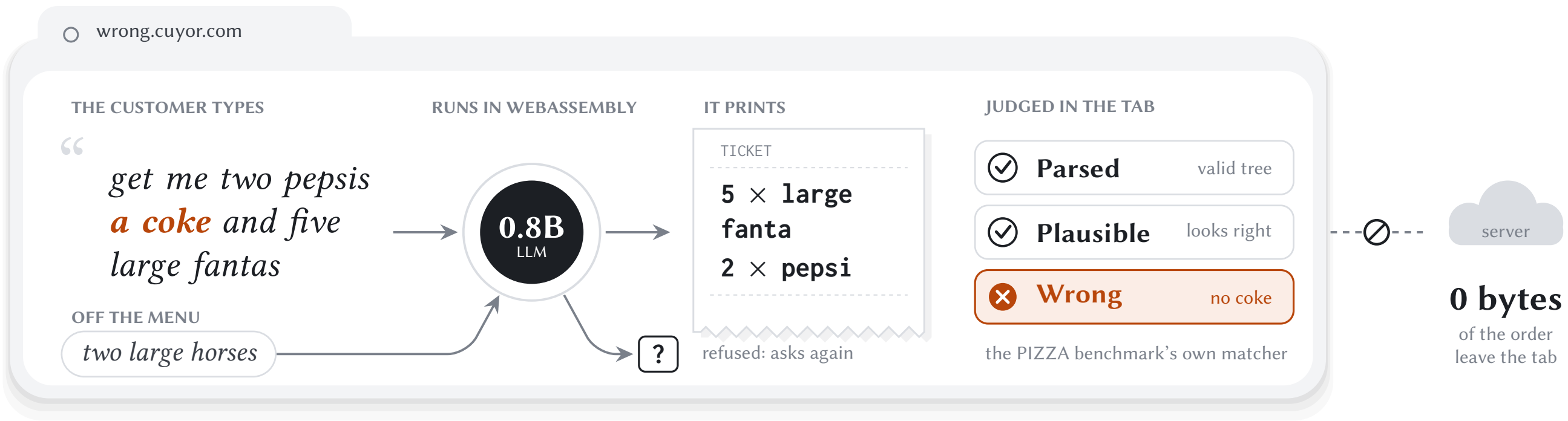
\begin{teaserfigure}
  \centering
  \resizebox{\textwidth}{!}{\definecolor{tkInk}{HTML}{1C1F24}
\definecolor{tkText}{HTML}{3A3F47}
\definecolor{tkSoft}{HTML}{7A808A}
\definecolor{tkLine}{HTML}{DADDE2}
\definecolor{tkWash}{HTML}{F3F4F6}
\definecolor{tkErr}{HTML}{B8460B}
\definecolor{tkErrWash}{HTML}{FBEDE6}
\begin{tikzpicture}[
  x=1cm, y=1cm, font=\sffamily,
  kicker/.style={font=\sffamily\bfseries\fontsize{5.4}{6}\selectfont, text=tkSoft, anchor=west},
  note/.style={font=\sffamily\fontsize{6}{7}\selectfont, text=tkSoft},
  verdict/.style={font=\sffamily\bfseries\fontsize{8.5}{9}\selectfont, anchor=west},
  arrow/.style={-{Stealth[length=1.9mm,width=1.7mm]}, line width=.7pt, draw=tkSoft},
]
\foreach \o/\a in {0.24/.03, 0.14/.045, 0.06/.06}
  \fill[tkInk, opacity=\a, rounded corners=12pt] (-\o/3,-\o) rectangle (14.25+\o/3,3.90);
\fill[tkWash, rounded corners=6pt] (0.30,3.55) rectangle (3.70,4.25);
\fill[tkWash, rounded corners=12pt] (0,0) rectangle (14.25,3.92);
\fill[white, rounded corners=9pt] (0.14,0.14) rectangle (14.11,3.55);
\draw[tkSoft, line width=.6pt] (0.66,3.94) circle (0.075);
\node[anchor=west, font=\sffamily\fontsize{6.4}{7}\selectfont, text=tkText] at (0.82,3.94) {wrong.cuyor.com};

\node[kicker] at (0.55,3.15) {THE CUSTOMER TYPES};
\node[font=\rmfamily\fontsize{24}{24}\selectfont, text=tkLine, anchor=north west]
  at (0.40,2.95) {\textquotedblleft};
\node[anchor=north west, text width=3.1cm, align=left,
      font=\rmfamily\itshape\fontsize{11}{13.4}\selectfont, text=tkInk]
  at (0.95,2.62)
  {get me two pepsis \textcolor{tkErr}{\textbf{a coke}} and five large fantas};

\node[kicker] at (4.45,3.15) {RUNS IN WEBASSEMBLY};
\draw[arrow] (4.15,1.80) -- (4.85,1.80);
\draw[tkLine, line width=.7pt] (5.65,1.80) circle (0.74);
\fill[tkInk] (5.65,1.80) circle (0.56);
\node[font=\sffamily\bfseries\fontsize{10.5}{11}\selectfont, text=white] at (5.65,1.86) {0.8B};
\node[font=\sffamily\fontsize{5}{5.5}\selectfont, text=white] at (5.65,1.59) {LLM};
\draw[arrow] (6.45,1.80) -- (7.15,1.80);

\node[kicker] at (0.55,0.88) {OFF THE MENU};
\node[anchor=west, draw=tkLine, line width=.5pt, rounded corners=7pt, inner xsep=6pt, inner ysep=3pt,
      font=\rmfamily\itshape\fontsize{8.5}{9}\selectfont, text=tkText]
  (horse) at (0.55,0.48) {two large horses};
\draw[arrow, rounded corners=6pt] (horse.east) -- (4.95,0.48) -- (5.30,1.12);
\draw[arrow, rounded corners=6pt] (6.00,1.12) -- (6.35,0.48) -- (6.75,0.48);
\node[anchor=west, draw=tkInk, line width=.6pt, rounded corners=2pt, inner sep=2.5pt,
      font=\ttfamily\bfseries\fontsize{8.5}{9}\selectfont, text=tkInk, minimum width=.42cm]
  (q) at (6.80,0.48) {?};
\node[note, anchor=west] at ([xshift=4pt]q.east) {refused: asks again};

\node[kicker] at (7.40,3.15) {IT PRINTS};
\fill[tkInk, opacity=.06] (7.48,2.86) -- (9.98,2.86) -- (9.98,0.66)
  decorate[decoration={zigzag, segment length=2.2mm, amplitude=.55mm}]{-- (7.48,0.66)} -- cycle;
\filldraw[fill=white, draw=tkLine, line width=.55pt] (7.40,2.92) -- (9.90,2.92) -- (9.90,0.72)
  decorate[decoration={zigzag, segment length=2.2mm, amplitude=.55mm}]{-- (7.40,0.72)} -- cycle;
\node[anchor=west, font=\ttfamily\fontsize{5.3}{6}\selectfont, text=tkSoft] at (7.58,2.68) {TICKET};
\draw[tkLine, dash pattern=on 1.2pt off 1.2pt] (7.58,2.50) -- (9.72,2.50);
\node[anchor=west, font=\ttfamily\bfseries\fontsize{8.8}{9}\selectfont, text=tkInk] at (7.58,2.20) {5 $\times$ large};
\node[anchor=west, font=\ttfamily\bfseries\fontsize{8.8}{9}\selectfont, text=tkInk] at (7.58,1.84) {fanta};
\node[anchor=west, font=\ttfamily\bfseries\fontsize{8.8}{9}\selectfont, text=tkInk] at (7.58,1.40) {2 $\times$ pepsi};
\draw[tkLine, dash pattern=on 1.2pt off 1.2pt] (7.58,1.12) -- (9.72,1.12);

\node[kicker] at (10.45,3.15) {JUDGED IN THE TAB};
\newcommand{\tkrow}[3]{%
  \draw[tkLine, line width=.5pt, rounded corners=4pt] (10.45,#1-.29) rectangle (13.90,#1+.29);
  \draw[tkInk, line width=.7pt] (10.75,#1) circle (0.15);
  \draw[tkInk, line width=.9pt, line cap=round, line join=round]
    (10.68,#1+.005) -- (10.735,#1-.055) -- (10.83,#1+.065);
  \node[verdict, text=tkInk] at (11.00,#1) {#2};
  \node[note, anchor=east] at (13.78,#1) {#3};
}
\tkrow{2.50}{Parsed}{valid tree}
\tkrow{1.80}{Plausible}{looks right}
\filldraw[fill=tkErrWash, draw=tkErr, line width=.6pt, rounded corners=4pt] (10.45,1.10-.29) rectangle (13.90,1.10+.29);
\fill[tkErr] (10.75,1.10) circle (0.15);
\draw[white, line width=.9pt, line cap=round]
  (10.69,1.04) -- (10.81,1.16) (10.69,1.16) -- (10.81,1.04);
\node[verdict, text=tkErr] at (11.00,1.10) {Wrong};
\node[note, anchor=east, text=tkErr] at (13.78,1.10) {no coke};
\node[note, anchor=west] at (10.45,0.50) {the PIZZA benchmark's own matcher};

\draw[tkSoft, line width=.7pt, dash pattern=on 2pt off 1.8pt] (14.30,1.80) -- (15.30,1.80);
\fill[white] (14.80,1.80) circle (0.15);
\draw[tkInk, line width=.8pt] (14.80,1.80) circle (0.13);
\draw[tkInk, line width=.8pt] (14.71,1.71) -- (14.89,1.89);
\begin{scope}[shift={(16.20,1.90)}, tkLine]
  \fill (-0.40,0.00) circle (0.22);
  \fill (0.02,0.14) circle (0.32);
  \fill (0.40,0.00) circle (0.22);
  \fill[rounded corners=4pt] (-0.62,-0.24) rectangle (0.62,0.06);
\end{scope}
\node[note] at (16.20,1.79) {server};
\node[font=\sffamily\bfseries\fontsize{11}{12}\selectfont, text=tkInk] at (16.20,1.05) {0 bytes};
\node[note, align=center] at (16.20,0.58) {of the order\\leave the tab};
\end{tikzpicture}
}
  \caption{A clean ticket for a wrong order. A real PIZZA test order goes through the
  demo: a 0.8B model running in the browser tab prints a ticket that parses and
  looks right, and the benchmark's own matcher, also in the tab, finds that the
  coke is missing. Input from off the menu is refused with ``\code{?}'', a
  detectable error, so the system asks again. No order text leaves the device.}
  \Description{A simplified browser tab, left to right. A quote reads ``get me
  two pepsis a coke and five large fantas'', with ``a coke'' emphasised. An arrow leads
  to a dark circle labelled 0.8B LLM, then to a receipt reading ``5 x large
  fanta, 2 x pepsi''. Three rows judge it: Parsed (check mark), Plausible (check
  mark), and Wrong (cross, highlighted), noting there is no coke. Below,
  a second input, ``two large horses'', goes through the same model and comes out
  as a question mark, refused so the system asks again. Outside
  the tab, a blocked dashed line leads to a grey server cloud, above the text
  ``0 bytes of the order leave the tab''.}
  \label{fig:teaser}
\end{teaserfigure}

\maketitle

\section{The ticket that prints perfectly}

Picture a pizza counter at rush hour. An order comes in, a small model turns it
into a ticket, and the ticket prints: \code{pizza 2 medium extra cheese}. It is
neat, it parses, and the kitchen starts cooking. The customer asked for
something else. Nothing crashed, nothing was logged, and nobody noticed until
the box was opened.

Our companion study~\cite{ahmed2026} calls this a \emph{silent error}: output
that parses cleanly but is wrong. Its opposite, a \emph{detectable} error, does
not parse, so the system can reject it and ask again. A grammar or a schema check
catches only the second kind. That study asks how the output format of a small
model shapes this split. This paper asks a different question: \textbf{what
changes when such a model leaves the server and runs in the visitor's browser
tab?} The two papers do not share results: the format study, its models and its
analyses belong to~\cite{ahmed2026}, which we cite only for the definitions above
and the reference row of Table~\ref{tab:acc}. Everything else here (the system,
its measurements in the browser and the lessons from deploying it) is new.

Running in the tab means order text never leaves the device and each inference
costs the site owner nothing. It also means there is no server log to audit, so
the person in front of the screen is the last check. We therefore built the page
as both a working parser and a place to look for its mistakes. The demo
contributes:
\begin{itemize}
  \item A fully client-side structured parser: a fine-tuned 0.8B model in the
        browser through WebAssembly, with no order text sent over the network
        (\S\ref{sec:system}).
  \item The benchmark's official matcher ported to TypeScript and checked
        against the Python original, so every ticket is judged in the tab with
        the benchmark's own rule.
  \item The study's learned guardrail~\cite{ahmed2026} put to work on the open
        Web: off-menu input gets ``\code{?}'', so the page can be left open to the
        public.
  \item Hands-on instruments to see, hunt and provoke silent errors on one's own
        device (\S\ref{sec:walk}).
  \item Evidence that the tab runs the model that was evaluated: on all \nTest{}
        test orders the browser writes the server's output for \srvAgree{} of
        them, and latency on consumer laptops (\S\ref{sec:eval}).
  \item Five lessons for shipping a fine-tuned model to the browser
        (\S\ref{sec:lessons}).
\end{itemize}

\section{Related work}

Client-side inference for the Web has moved from small vision models to
language models, through WebAssembly builds of llama.cpp such as
wllama~\cite{wllama,llamacpp} and WebGPU runtimes such as WebLLM~\cite{webllm}
and Transformers.js~\cite{transformersjs}. These runtimes are usually shown with
general chat models. We instead ship a narrow, fine-tuned model whose every
output can be scored exactly in the page. Structured output is often made safe
by constraining decoding to a grammar~\cite{gcd,outlines}. A grammar guarantees
that the output parses, which removes detectable errors but not silent ones.
Interactive explanations of model behaviour on the Web~\cite{tfplayground}
inspired our instruments, and human-in-the-loop adversarial data collection, as
in Dynabench~\cite{dynabench}, inspired our ``break it'' mode. Here it runs with
no server in the loop. Measurements of in-browser inference find it several times
slower than native code and costly in memory~\cite{wang2024anatomizing}, and newer
WebGPU backends for llama.cpp reduce that memory cost~\cite{levine2026llamaweb}.
We measure a different property: whether the model in the tab answers like the
model that was evaluated, and whether it is fast enough to type against.

\section{A kitchen in a browser tab}
\label{sec:system}

\begin{figure}[t]
  \centering
  \resizebox{\columnwidth}{!}{\definecolor{aInk}{HTML}{1C1F24}
\definecolor{aText}{HTML}{3A3F47}
\definecolor{aSoft}{HTML}{7A808A}
\definecolor{aLine}{HTML}{DADDE2}
\definecolor{aWash}{HTML}{F3F4F6}
\definecolor{aErr}{HTML}{B8460B}
\providecommand{\bx}[2]{{\bfseries\fontsize{6.4}{7}\selectfont\color{aInk}#1}\\[1pt]#2}
\begin{tikzpicture}[
  x=1cm, y=1cm, font=\sffamily,
  box/.style={draw=aLine, fill=white, line width=.5pt, rounded corners=3pt,
              minimum width=1.62cm, minimum height=1.12cm, inner sep=2pt, align=center,
              font=\sffamily\fontsize{5.2}{6}\selectfont, text=aText},
  ttl/.style={font=\sffamily\bfseries\fontsize{6.4}{7}\selectfont, text=aInk},
  sub/.style={font=\sffamily\fontsize{5.2}{6}\selectfont, text=aText},
  flow/.style={-{Stealth[length=1.5mm,width=1.3mm]}, line width=.55pt, draw=aSoft},
  lab/.style={font=\sffamily\fontsize{4.9}{5.5}\selectfont, text=aSoft, fill=white, inner sep=.8pt},
]
\fill[aWash, rounded corners=6pt] (0,0) rectangle (6.30,3.55);
\node[anchor=north west, font=\sffamily\bfseries\fontsize{5.2}{6}\selectfont, text=aSoft]
  at (0.12,3.50) {BROWSER TAB, ON THE VISITOR'S DEVICE};

\node[box] (page) at (1.00,2.48) {\bx{Page}{live box, game,\\gate}};
\node[box] (wrk)  at (3.12,2.48) {\bx{Web Worker}{llama.cpp in\\WebAssembly}};
\node[box] (opfs) at (5.24,2.48) {\bx{OPFS cache}{q8\_0 model, URL\\keyed by SHA-256}};
\node[box] (score) at (1.00,0.80) {\bx{Scorer}{decoder + matcher,\\checked TS ports}};
\node[box] (sw)   at (3.12,0.80) {\bx{Service worker}{isolation headers,\\so: threads}};
\node[box] (bench) at (5.24,0.80) {\bx{Bench page}{same engine,\\JSON to download}};

\draw[flow] ([yshift=4pt]page.east) -- node[lab, above=1pt] {order} ([yshift=4pt]wrk.west);
\draw[flow] ([yshift=-4pt]wrk.west) -- node[lab, below=1pt] {ticket} ([yshift=-4pt]page.east);
\draw[flow] (opfs.west) -- node[lab, above=1pt] {weights} (wrk.east);
\draw[flow] (page.south) -- node[lab, right=1pt] {verdict} (score.north);
\draw[flow, dashed] (sw.north) -- (wrk.south);
\draw[flow] (bench.north) -- (opfs.south -| bench.north);

\node[box, minimum width=1.45cm, minimum height=.9cm] (hf) at (7.55,2.48) {\bx{Hugging Face}{model file}};
\draw[flow] (hf.west) -- node[lab, above=1pt, align=center] {once,\\834\,MB} (opfs.east);
\node[box, fill=aWash, minimum width=1.45cm, minimum height=.9cm] (srv) at (7.55,0.80) {\bx{Any server}{analytics, logs}};
\draw[aErr, line width=.8pt, dash pattern=on 2pt off 1.4pt] (6.30,0.80) -- (srv.west);
\fill[white] (6.55,0.80) circle (0.13);
\draw[aErr, line width=.8pt] (6.55,0.80) circle (0.11) (6.475,0.725) -- (6.625,0.875);
\node[lab, text=aErr, anchor=north, align=center, fill=none] at (6.55,0.55) {0 bytes\\of order\\text};
\end{tikzpicture}}
  \caption{What runs where. Everything in the grey box runs in the visitor's
  tab. The model file arrives once from Hugging Face; no order text leaves.}
  \Description{A diagram. Inside a grey box labelled ``browser tab, on the
  visitor's device'': the page sends the order to a Web Worker running llama.cpp
  in WebAssembly and gets the ticket back; the worker reads the model from an
  OPFS cache keyed by its SHA-256; the page passes tickets to a scorer (decoder
  and matcher); a service worker supplies the headers that allow threads; a bench
  page uses the same engine. Outside the box, Hugging Face sends the model file
  once, and a dashed line to ``any server'' is blocked, marked ``0 bytes of order
  text''.}
  \label{fig:arch}
\end{figure}
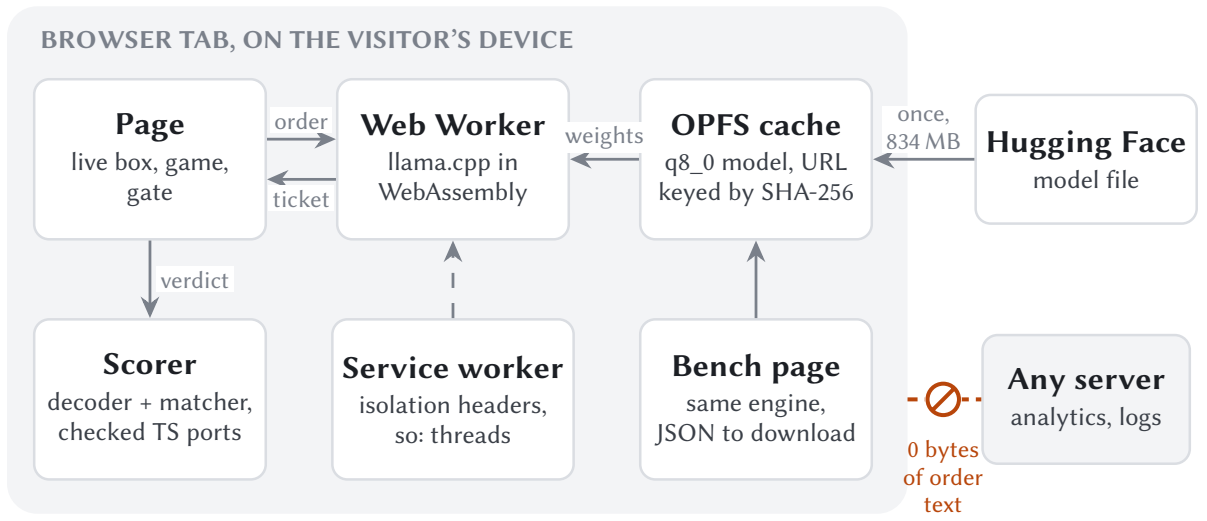

Fig.~\ref{fig:arch} shows the parts. \textbf{Model.} The model is Qwen3.5-0.8B~\cite{qwen} %
with a LoRA adapter~\cite{lora}, fine-tuned on PIZZA~\cite{pizza} to write the
target tree in plain words (for example \code{pizza 2 medium extra cheese}), the
encoding recommended by the companion study~\cite{ahmed2026}. It is seed~0 of
that study's word-format model, so like all its models it is also trained on
off-menu input labelled ``\code{?}'' (\negPct{} of examples) and on spoken
self-corrections (\corrPct). The merged weights are quantized to 8 bits (GGUF
\code{q8\_0}, \modelMB\,MB) and published on Hugging Face with their SHA-256;
every measurement below first checks the local file against it.

\textbf{Runtime.} The model is not downloaded until the visitor asks for it.
Before that, the page replays saved model outputs, so every instrument works
without the download. The page then loads the model with wllama~\cite{wllama}, a
WebAssembly build of llama.cpp~\cite{llamacpp}, which runs inference in a Web
Worker so the page stays responsive. Multithreaded WebAssembly needs
cross-origin isolation. A static host cannot set the required headers, so a small
service worker adds them. The model file is stored in the Origin Private File
System (OPFS) by wllama's cache, so later visits load it from disk; if storage is
refused, as in some private windows, the page downloads it into memory instead.
Decoding is greedy, with the prompt template, stop string and \tokenCap-token cap
read from the same export manifest used to verify the GGUF on the server, so the
browser asks the model exactly what the evaluation asked it.

\textbf{Scoring in the tab.} PIZZA judges a parse with an unordered tree match
written in Python. We ported it to TypeScript and, rather than trusting the port,
compared it with the official matcher on \parityN{} saved predictions, with
\parityDisagree{} disagreements. The matcher
compares trees, so the page also decodes each ticket back to a tree with a port
of the reference decoder. On \decOutputs{} saved outputs of the study's word-format
models it has \decDisagree{} disagreements with the Python decoder.
The page can therefore mark a ticket correct, detectable or silent with the
benchmark's own rule.

\textbf{Guardrail.} A public page receives anything: greetings, gibberish, emoji,
``two large horses''. Rather than a separate filter, the model itself answers
``\code{?}'' outside the menu. The companion study measures this guardrail on
held-out data~\cite{ahmed2026}; on the page we check that it survives
quantization and the browser: of the \probeN{} off-menu probes that the page's
gate animation replays, \probeRefused{} get ``\code{?}'' in the tab, and
Table~\ref{tab:acc} reports how often it refuses a real order. A refusal is a detectable error by
design: the system can ask again.

\textbf{Privacy by construction.} Inference, scoring and game state all stay in
the tab, and the page has no analytics. %
We recorded all network traffic in Chromium while typing \netOrders{} test orders (\netChars{}
characters) key by key into the live box after the model had loaded: the page
made \netRequests{} network requests (\netBytes{} bytes sent).
Before that point, the only third-party requests are two web fonts from Google
Fonts and, on the live site, the one-time model download from Hugging Face; none
carries order text. %

\section{What a visitor does}
\label{sec:walk}

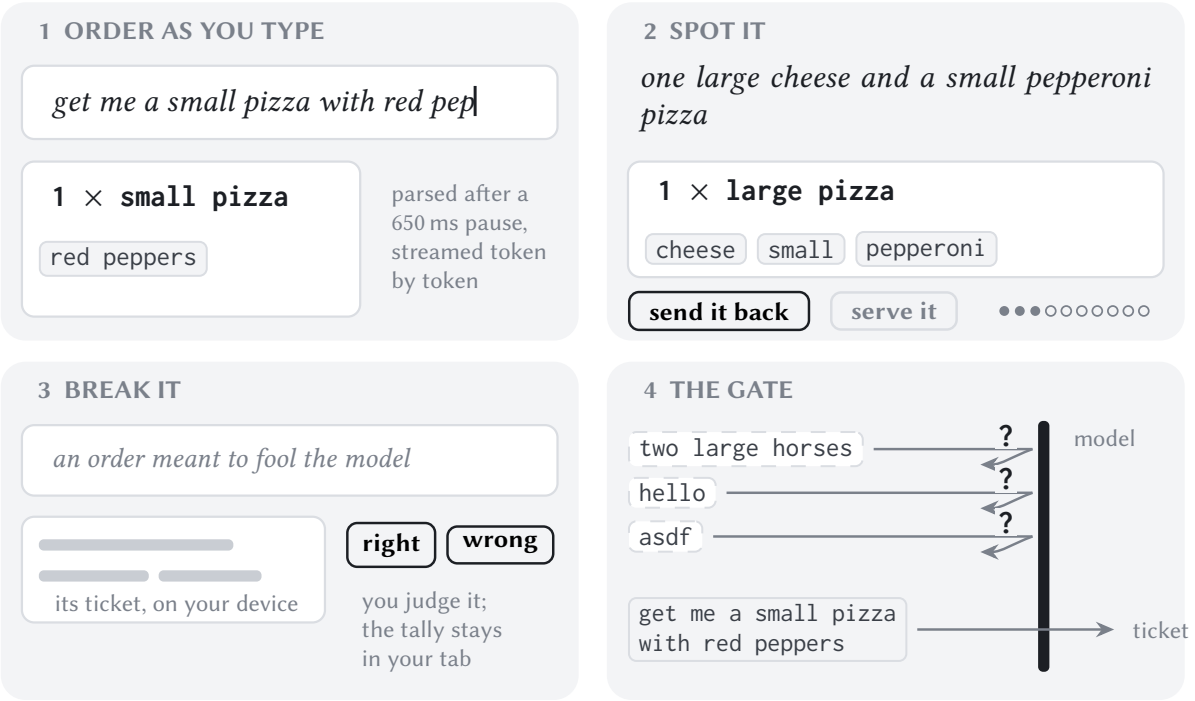
\begin{figure}[t]
  \centering
  \resizebox{\columnwidth}{!}{\definecolor{vInk}{HTML}{1C1F24}
\definecolor{vText}{HTML}{3A3F47}
\definecolor{vSoft}{HTML}{7A808A}
\definecolor{vLine}{HTML}{DADDE2}
\definecolor{vWash}{HTML}{F3F4F6}
\definecolor{vBar}{HTML}{C9CDD3}
\begin{tikzpicture}[
  x=1cm, y=1cm, font=\sffamily,
  kick/.style={font=\sffamily\bfseries\fontsize{5.2}{6}\selectfont, text=vSoft, anchor=west},
  note/.style={font=\sffamily\fontsize{4.9}{5.8}\selectfont, text=vSoft},
  say/.style={font=\rmfamily\itshape\fontsize{6.4}{7.6}\selectfont, text=vInk},
  head/.style={font=\ttfamily\bfseries\fontsize{6.2}{7}\selectfont, text=vInk, anchor=west},
  chip/.style={font=\ttfamily\fontsize{5.4}{6}\selectfont, text=vText, draw=vLine, fill=vWash,
               rounded corners=1.5pt, inner xsep=2pt, inner ysep=1.2pt, anchor=west},
  btn/.style={font=\sffamily\bfseries\fontsize{5.2}{6}\selectfont, draw=vInk, line width=.5pt,
              rounded corners=2pt, inner xsep=4pt, inner ysep=2pt, anchor=west},
  box/.style={draw=vLine, fill=white, line width=.45pt, rounded corners=2.5pt},
  flow/.style={-{Stealth[length=1.3mm,width=1.1mm]}, line width=.5pt, draw=vSoft},
]
\foreach \x/\y in {0/2.55, 4.3/2.55, 0/0, 4.3/0}
  \fill[vWash, rounded corners=5pt] (\x,\y) rectangle ++(4.1,2.4);

\node[kick] at (0.15,4.75) {1\enspace ORDER AS YOU TYPE};
\draw[box] (0.15,3.98) rectangle (3.95,4.5);
\node[say, anchor=west] at (0.25,4.24) {get me a small pizza with red pep\textcolor{vInk}{\rule[-.15em]{.6pt}{.95em}}};
\draw[box] (0.15,2.72) rectangle (2.55,3.82);
\node[head] at (0.25,3.56) {1 $\times$ small pizza};
\node[chip] at (0.27,3.13) {red peppers};
\node[note, align=left, anchor=west] at (2.66,3.27) {parsed after a\\650\,ms pause,\\streamed token\\by token};

\node[kick] at (4.45,4.75) {2\enspace SPOT IT};
\node[say, align=left, anchor=north west, text width=3.75cm] at (4.42,4.62)
  {one large cheese and a small pepperoni pizza};
\draw[box] (4.45,3.0) rectangle (8.25,3.82);
\node[head] at (4.55,3.6) {1 $\times$ large pizza};
\node[chip] (c1) at (4.57,3.2) {cheese};
\node[chip] (c2) at ([xshift=2pt]c1.east) {small};
\node[chip] at ([xshift=2pt]c2.east) {pepperoni};
\node[btn] (back) at (4.45,2.76) {send it back};
\node[btn, draw=vLine, text=vSoft] at ([xshift=4pt]back.east) {serve it};
\foreach \i in {0,...,9} {
  \ifnum\i<3 \fill[vSoft] (7.12+\i*0.11,2.76) circle (0.035);
  \else \draw[vSoft, line width=.35pt] (7.12+\i*0.11,2.76) circle (0.035); \fi
}

\node[kick] at (0.15,2.2) {3\enspace BREAK IT};
\draw[box] (0.15,1.45) rectangle (3.95,1.95);
\node[note, anchor=west, font=\rmfamily\itshape\fontsize{5.6}{6}\selectfont] at (0.25,1.7) {an order meant to fool the model};
\draw[box] (0.15,0.55) rectangle (2.3,1.3);
\fill[vBar, rounded corners=1pt] (0.27,1.06) rectangle (1.65,1.14);
\fill[vBar, rounded corners=1pt] (0.27,0.84) rectangle (1.05,0.92);
\fill[vBar, rounded corners=1pt] (1.12,0.84) rectangle (1.85,0.92);
\node[note, anchor=west] at (0.27,0.67) {its ticket, on your device};
\node[btn, inner xsep=3pt] (r) at (2.45,1.1) {right};
\node[btn, inner xsep=3pt] at ([xshift=2pt]r.east) {wrong};
\node[note, align=left, anchor=north west] at (2.45,0.88) {you judge it;\\the tally stays\\in your tab};

\node[kick] at (4.45,2.2) {4\enspace THE GATE};
\fill[vInk, rounded corners=1pt] (7.36,0.2) rectangle (7.44,1.98);
\node[note, anchor=west] at (7.5,1.86) {model};
\foreach \t/\yy in {two large horses/1.78, hello/1.47, asdf/1.16} {
  \node[chip, fill=white, dashed] (q) at (4.45,\yy) {\t};
  \draw[vSoft, line width=.5pt] ([xshift=2pt]q.east) -- (7.32,\yy);
  \draw[flow] (7.32,\yy) to[out=200, in=0] (6.95,\yy-0.11);
  \node[font=\ttfamily\bfseries\fontsize{6.4}{6}\selectfont, text=vInk, fill=vWash, inner sep=.6pt] at (7.13,\yy+0.1) {?};
}
\node[chip, align=left, font=\ttfamily\fontsize{5.2}{6}\selectfont] (ok) at (4.45,0.5)
  {get me a small pizza\\with red peppers};
\draw[flow] ([xshift=2pt]ok.east) -- (7.9,0.5);
\node[note, anchor=west] at (7.92,0.5) {ticket};
\end{tikzpicture}}
  \caption{What a visitor does. (1) The ticket for a real test order streams in as
  they type. (2) A saved ticket that parses and is wrong: two pizzas became one,
  with ``small'' as a topping. (3) Their own attempt to fool the model, judged by
  them. (4) Off-menu input bounces off with ``\code{?}''; an order passes.}
  \Description{Four panels. Panel 1, order as you type: an input reads ``get me a
  small pizza with red pep'' and a ticket shows ``1 x small pizza, red peppers''.
  Panel 2, spot it: the order ``one large cheese and a small pepperoni pizza'' and a
  ticket ``1 x large pizza'' with chips cheese, small, pepperoni, above buttons
  ``send it back'' and ``serve it'' and ten progress dots. Panel 3, break it: an
  input box, a ticket drawn as grey bars, and buttons ``right'' and ``wrong''.
  Panel 4, the gate: ``two large horses'', ``hello'' and ``asdf'' bounce off a bar
  labelled model with question marks, while ``get me a small pizza with red
  peppers'' passes through to a ticket.}
  \label{fig:visitor}
\end{figure}
The page reads top to bottom as a short story: a scroll-driven scene shows one
silent error being made, then the visitor takes over (Fig.~\ref{fig:visitor}).

\textbf{1. Order as you type.} The visitor types an order into the live box. When
they pause for 650\,ms, or press Enter, %
the page sends the current text to the model, aborting any parse still running
for older text, and streams the ticket in token by token. A fast typist thus
always sees the parse of their latest text, never a queue of stale ones.

\textbf{2. Spot it.} The visitor plays 10 rounds, %
each a real ticket that the deployed model wrote for a real test order in its
server evaluation, and must send back the wrong ones. Four rounds are silent %
errors. Every one parses, so the game ends by pointing out that a parser would
have caught none of them.

\textbf{3. Break it.} The visitor writes an order meant to fool the live model on
their own device, reads the ticket, and marks it right or wrong. The page keeps a
local tally of breaks; nothing is submitted. %

\textbf{4. The gate.} An animation sends real inputs at the model: off-menu ones
bounce off with ``\code{?}'', menu orders pass through and print. It shows the
guardrail at work on the probes from its evaluation.

\begin{figure}[t]
  \centering
  \resizebox{\columnwidth}{!}{\input{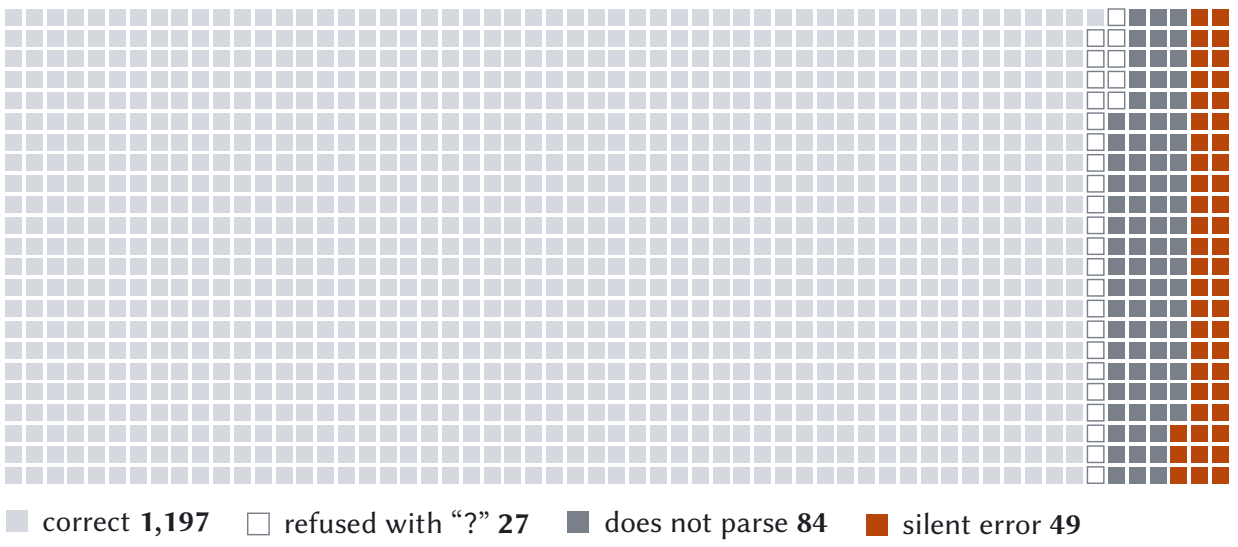}}
  \caption{All \nTest{} PIZZA test orders, each a tile, run by the deployed model
  inside Chromium and scored in the tab. Tiles are grouped by outcome. The dark
  tiles on the right parse cleanly and are wrong.}
  \Description{A rectangle of 59 by 23 small squares. Most are light grey
  (\brCorrect{} correct). Then come \brRefused{} outlined squares (refused with a
  question mark), \brMalformed{} mid-grey squares (output that does not parse),
  and \brSilentN{} dark squares (silent errors).}
  \label{fig:grid}
\end{figure}

\section{Does the tab keep its promises?}
\label{sec:eval}

A demo is only useful if the model in the tab is the model that was evaluated,
and if it is fast enough to use while typing. We test both.

\textbf{Same model, same answers.} Quantization and a WebAssembly runtime could
both change outputs. Before release, the \code{q8\_0} GGUF agreed with the saved
full-precision outputs on \verifyAgree{} of a \verifyN-order check on the server.
In the browser we run all \nTest{} test orders through the page's own engine,
driven by Playwright in Chromium, and score them with the in-tab matcher
(Table~\ref{tab:acc}, Fig.~\ref{fig:grid}).

In the tab, the deployed model gets \brCorrect{} of \nTest{} orders exactly
right. Of its \brDetectable{} detectable errors, \brRefused{} are refusals with
``\code{?}'' and \brMalformed{} do not parse.
Only \brSilentN{} orders (\brSilentPct) are silent errors, and those are what
the page is built to expose. The median parse took \brMedianS{} on an Apple M1
in headless Chromium. \brRestartNote

The browser and the server agree on exact match, but that alone could hide
changes that cancel out, so we compare every output. \fidelityReading

\begin{table}[t]
  \caption{Where the \nTest{} PIZZA test orders go: exact match, silent error,
  refused with ``\code{?}'', or output that does not parse (rows sum to 1). The
  cited row is the companion study's word-format model~\cite{ahmed2026}, mean of
  \citedSeeds{} seeds at full precision; the deployed model is its seed~0,
  quantized and run in the browser.}
  \label{tab:acc}
  \small
  \setlength{\tabcolsep}{4pt}
  \begin{tabular*}{\columnwidth}{@{}l@{\extracolsep{\fill}}cccc@{}}
    \toprule
    Setting & Exact & Silent & Refused & No parse \\
    \midrule
    Cited, \citedSeeds{} seeds~\cite{ahmed2026} & \citedEm & \citedSilent & \citedRefused & \citedNoParse \\
    Deployed, browser & \brEm & \brSilent & \brRefusedRate & \brNoParse \\
    \midrule
    \multicolumn{5}{l}{Browser output identical to the server's, order by order: \srvAgree} \\
    \bottomrule
  \end{tabular*}
\end{table}

\textbf{Fast enough to use?} We do not claim real-time guarantees. We use the
classic limits from human-computer interaction: about 0.1\,s feels instantaneous
and about 1\,s keeps the user's flow of thought~\cite{card1991,nielsen1993}. A
hidden benchmark page on the live site runs \benchTimed{} fixed test orders through
the real engine after \benchWarmup{} warm-up orders, and records the time from handing the
text to the model until the final ticket is drawn. The 650\,ms typing pause is
not included. It also reports the one-time cost of a first visit, and it
uploads nothing; the visitor downloads the result. Table~\ref{tab:lat} reports
the median, p90 and share of orders under each limit.

\begin{table}[t]
  \caption{In-browser latency per order, from submitting the text to the
  finished ticket, share of orders under each limit, and first-visit load
  time (cold cache). Model: \code{q8\_0}, \modelMB\,MB. $^w$: warm cache.}
  \label{tab:lat}
  \small
  \setlength{\tabcolsep}{3pt}
  \begin{tabular*}{\columnwidth}{@{}l@{\extracolsep{\fill}}ccccc@{}}
    \toprule
    Device, browser & Median & p90 & $<$0.1\,s & $<$1\,s & Load \\
    \midrule
    \latencyRows
    \bottomrule
  \end{tabular*}
\end{table}

\latencyReading{} \latencyBreakdown

\section{Lessons from shipping to the tab}
\label{sec:lessons}

Each of these cost us a wrong number or a broken page before we caught it.

\textbf{1. Start every parse from a clean state.} The runtime embeds llama.cpp's
server in the worker, which by default reuses the previous prompt's cache. With our
hybrid model, whose layers keep a recurrent state, this crashed the worker at a
fixed point in a fixed sequence of orders. We turn the prompt cache off, which
also makes each parse independent of the last, as in the server evaluation. The
worker still fails rarely (\S\ref{sec:eval}), so the page reloads it from the
cache and retries the order once, and benchmarks report those orders apart.

\textbf{2. Key the cache by content, not by name.} The model is cached by its
URL. A new model published under the same name was served from the old cache,
while the page reported the new one. The model's SHA-256 is now part of the URL.

\textbf{3. Measure only in a secure context.} The file cache and multithreading
need HTTPS or \code{localhost}. Opened over plain HTTP on a local network, the
same page runs on one thread and cannot cache the model, so the benchmark page
flags such runs and the paper uses none.

\textbf{4. Compare outputs, not scores.} Identical exact match hid
\srvDiffN{} changed outputs (\S\ref{sec:eval}). Only an order-by-order comparison
shows whether the tab runs the evaluated model.

\textbf{5. Make the download optional.} A first visit costs more than any parse
(Table~\ref{tab:lat}), so the page works from saved outputs and downloads the
model only when asked.

\section{Demonstration plan}

At the venue, visitors use the live page on laptops we bring, with the model
already cached so the demo needs no network. Visitors with their own laptop can
open the same page; the \modelMB\,MB model downloads once and then runs offline.
A poster shows the ticket rail, and the break-it tally on the laptop shows how
many silent errors visitors caused.
Code: \url{https://github.com/UmarGit/coffee-llm}. Model:
\url{https://huggingface.co/umarsheikh303/pizza-words-gguf}.

\section{Ethics and data use}

\textbf{Data.} PIZZA~\cite{pizza} is released under CC-BY-NC-4.0. We use it for
research only, and the released model inherits the non-commercial license.
\textbf{Visitors.} The page sends no visitor text anywhere, has no analytics, and
keeps game results in the tab. Visitors who never press the load button
download no model.
\textbf{Limits.} The model is a research artifact trained on pizza orders. It is
not meant for real ordering, and a silent error on this page costs nothing,
which is why it is a safe place to look for one.

\section{Conclusion}

A clean parse is not a correct one, and when the model runs in the visitor's tab
there may be no one else to notice. Our demo puts the parser, its scorer and its
failure modes in one page that runs on the visitor's own device and sends no
order text anywhere. A quantized 0.8B model in a browser tab reproduces its
server evaluation, writing the same output for \srvAgree{} of the test orders,
but only an order-by-order check shows it. The page, the benchmark page and the
checked ports of the scorer are open source, as a pattern for private,
client-side structured extraction on the Web.

\bibliographystyle{ACM-Reference-Format}
\bibliography{refs}

\end{document}